\documentclass{llncs}

\usepackage{t1enc}
\usepackage[latin2]{inputenc}
\usepackage[english]{babel}
\usepackage{url}
\usepackage{amsmath}
\usepackage[mathscr]{eucal}
\usepackage{subfig}
\usepackage{bm}
\usepackage{ntheorem}
\usepackage{amssymb}
\usepackage{graphicx}

\renewcommand{\b}{\mathbf}
\newcommand{\R}{\mathbb{R}}
\newcommand{\N}{\mathcal{N}}
\newcommand{\T}{\mathcal{T}}
\newcommand{\IW}{\mathcal{IW}}

\newtheorem{thm}{Theorem}[section]
\newtheorem{cor}[thm]{Corrolary}
\newtheorem{lem}[thm]{Lemma}
\begin{document}

\title{D-optimal Bayesian Interrogation for Parameter and Noise Identification of Recurrent Neural Networks}

\author{Barnab{\'a}s P{\'o}czos \thanks{Present address: Department of Computing Science, University of Alberta,
 Athabasca Hall, Edmonton, Canada, T6G 2E8}
        and Andr{\'a}s L{\H{o}}rincz}

\institute{Department of Information Systems, E\"{o}tv\"{o}s Lor{\'a}nd University,\\
              P{\'a}zm{\'a}ny P. s{\'e}t{\'a}ny 1/C, Budapest H-1117, Hungary\\
              WWW home page: \url{http://nipg.info}\\
              \email{poczos@cs.ualberta.ca, lorincz@inf.elte.hu}}
\maketitle
\begin{abstract}%   <- trailing '%' for backward compatibility of .sty file
We introduce a novel online Bayesian method for the identification of a family of
noisy recurrent neural networks (RNNs). We develop Bayesian active learning
technique in order to optimize the interrogating stimuli given past experiences. In
particular, we consider the unknown parameters as stochastic variables and use the
D-optimality principle, also known as `\emph{infomax method}', to choose optimal
stimuli. We apply a greedy technique to maximize the information gain concerning
network parameters at each time step. We also derive the D-optimal estimation of the
additive noise that perturbs the dynamical system of the RNN. Our analytical results
are approximation-free. The analytic derivation gives rise to attractive quadratic
update rules.
\end{abstract}

%\begin{keywords}
%Active learning, system identification, online Bayesian learning, D-optimality, infomax control,
%optimal design
%\end{keywords}

\section{Introduction}\label{sec:introduction}
When studying online systems it is of high relevance to facilitate fast information
gain concerning the system \cite{fedorov72theory,cohn94neural}. As an example,
consider the research on real neurons. In one of the experimental paradigms,
researchers look for the stimulus that maximizes the response of the neuron
\cite{deCharmes98optimizing,foldiak01stimulus}. Another approach searches for
stimulus distribution that maximizes mutual information between stimulus and
response \cite{machens05testing}. A recent technique assumes that the unknown system
belongs to the family of generalized linear models \cite{lewi07realtime} and treats
the parameters as probabilistic variables. Then the goal is to find the optimal
stimuli by maximizing mutual information between the parameter set and the response
of the system.

We are interested in the active learning
\cite{mackay92information,cohn96active,fukumizu00statistical,sugiyama06active} of
noisy recurrent artificial neural networks (RNNs), when we have the freedom to
interrogate the network and to measure the responses. Our framework is similar to
the generalized linear model (GLM) approach used by \cite{lewi07realtime}: we would
like to choose interrogating, or `\emph{control}' inputs in order to (i) identify
the parameters of the network and (ii) estimate the additive noise efficiently. From
now on, we use the terms \emph{control} and \emph{interrogation} interchangeably;
control is the conventional expression, whereas the word interrogation expresses our
aims better. We apply online Bayesian learning
\cite{opper99bayesian,solla98optimal,honkela03online,ghahramani00online} to
accomplish our task. For Bayesian methods prior updates often lead to intractable
posterior distributions such as a mixture of exponentially numerous distributions.
Here, we show that in our model computations are both tractable and
approximation-free. Further, the emerging learning rules are simple. We also show
that different stimuli are needed for the same RNN model depending on whether the
goal is to estimate the weights of the RNN or the additive noise that perturbs the
RNN. Hereafter we will refer to this noise as the `driving noise' of the RNN.

Our approach, which optimizes control online in order to gain maximum information
concerning the parameters, falls into the realm of Optimal Experimental Design, or
Optimal Bayesian Design
\cite{kiefer59optimum,fedorov72theory,steinberg84experimental,toman93robust,pukelsheim93optimal}.
Several optimality principles have been worked out and their efficiencies have been
studied extensively in the literature. For a review see, e.g.,
\cite{chaloner95bayesian}. Our approach corresponds to the so-called D-optimality
\cite{bernardo79expected,stone59application}, which is equivalent to the information
maximization (infomax) principle applied by \cite{lewi07realtime}. We use both
terms, the term \emph{D-optimality} and the term \emph{infomax}, to designate our
approach. To the best of our knowledge, D-optimality has not been applied to the
typical non-spiking stochastic artificial recurrent neural network model that we
treat here.

The contribution of this paper can be summarized as follows: We use the D-optimality
(infomax) principle and derive cost functions and algorithms for (i) the parameter
learning of the stochastic RNN and (ii) the estimation of its driving noise. We show
that, (iii) using the D-optimality interrogation technique, these two tasks are not
compatible with each other: greedy control signals derived from the D-optimality
principle for parameter estimation are suboptimal (basically the worst possible) for
the estimation of the driving noise and vice versa. We show that (iv) D-optimal cost
functions lead to simple greedy optimization rules both for the parameter estimation
and for the noise estimation, respectively. Investigation of non-greedy multiple
step optimizations, which may achieve more efficient estimation of the network
parameters and the noise, seems difficult and is beyond the scope of the present
paper. However, (v) for the task of estimating the driving noise we introduce a
non-greedy multiple step look-ahead heuristics.

The paper is structured as follows: In Section~\ref{s:Model} we introduce our model.
Section~\ref{s:Bayesian} concerns the Bayesian equations of the RNN model.
Section~\ref{s:Infomax} derives the optimal control for its parameter identification
starting from the D-optimality (infomax) principle. Section~\ref{s:Estimating} deals
with our second task, when the goal is the estimation of the driving noise of the
RNN. The paper ends with a short discussion and some conclusions
(Section~\ref{s:Discussion_and_Conclusions}).

%%%%%%%%%%%%%%%%%%%%%%%%%%%%%%%%%%%%%%%%%%%%%%%%
\section{The Model} \label{s:Model}
%%%%%%%%%%%%%%%%%%%%%%%%%%%%%%%%%%%%%%%%%%%%%%%%

We introduce our model here. Let $P(\b{e})=\N_{\b{e}}(\b{m},\b{V})$ denote the probability density
of a normally distributed stochastic variable $\b{e}$ with mean $\b{m}$ and covariance matrix
$\b{V}$. Let us assume that we have $d$ simple computational units called `\emph{neurons}' in a
recurrent neural network:
\begin{eqnarray}
\b{r}_{t+1}&=&g\left(\sum_{i=0}^I \b{F}_i\b{r}_{t-i}+\sum_{j=0}^J
\b{B}_j\b{u}_{t+1-j}+ \b{e}_{t+1}\right), \label{e:Modell}
\end{eqnarray}
where $\{\b{e}_t\}$, the driving noise of the RNN, denotes temporally independent
and identically distributed (i.i.d.) stochastic variables and
$P(\b{e}_t)=\N_{\b{e}_t}(\b{0},\b{V})$, $\b{r}_t \in \R^d$ represents the observed
activities of the neurons at time $t$. Let $\b{u}_t\in \R^c$ denote the control
signal at time $t$. The neural network is formed by the weighted delays represented
by matrices $\b{F}_i$ ($i=0, \ldots , I$) and $\b{B}_j$ ($j=0, \ldots , J$), which
connect neurons to each other and also the control components to the neurons,
respectively. Control can also be seen as the means of interrogation, or the
stimulus to the network \cite{lewi07realtime}. We assume that function $g: \R^d \to
\R^d$ in \eqref{e:Modell} is known and invertible. The computational units, the
neurons, sum up weighted previous neural activities as well as weighted control
inputs. These sums are then passed through identical non-linearities according to
Eq.~\eqref{e:Modell}. Our goal is to estimate the parameters $\b{F}_i\in \R^{d
\times d}$ ($i=0,\ldots,I$), $\b{B}_j \in \R^{d \times c}$ ($j=0,\ldots,J$) and the
covariance matrix $\b{V}$, as well as the driving noise $\b{e}_t$ by means of the
control signals.

In artificial neural network terms, \eqref{e:Modell} is in the form of \emph{rate code models}. In
our rate code model, noise, control, and the recurrent activities influence the firing rates
similarly. We show that analytic cost functions emerge for this model that are free of
approximation.

%%%%%%%%%%%%%%%%%%%%%%%%%%%%%%%%%%%%%%%%%%%%%%%%%%%
\section{Bayesian Approach} \label{s:Bayesian}
%%%%%%%%%%%%%%%%%%%%%%%%%%%%%%%%%%%%%%%%%%%%%%%%%%

Here we embed the estimation task into the Bayesian framework. First, we introduce the following
notations: \mbox{$\b{x}_{t+1}=[\b{r}_{t-I}; \ldots;\b{r}_{t};\b{u}_{t-J+1};\ldots;\b{u}_{t+1}]$},
\mbox{${\b{y}_{t+1}}= {g^{-1}(\b{r}_{t+1})}$},
\mbox{$\b{A}=[\b{F}_I,\ldots,\b{F}_0,\b{B}_{J},\ldots, \b{B}_{0}] \in \R^{d \times m}$}. With these
notations, model~\eqref{e:Modell} reduces to a linear equation
\begin{eqnarray}
\b{y}_{t}&=&\b{A}\b{x}_{t}+\b{e}_{t}. \label{e:simplified_model}
\end{eqnarray}
To fulfill our goal, the online  estimation of the unknown quantities (parameter
matrix $\b{A}$, noise $\b{e}_t$ and its covariance matrix $\b{V}$), we rely on
Bayes' method. We assume that prior knowledge is available and we update our
posteriori knowledge on the basis of the observations. Control will be chosen at
each instant to provide maximal expected information concerning the quantities we
have to estimate. Starting from an arbitrary prior distribution of the parameters
the posterior distribution needs to be computed. This can be highly complex,
however, so approximations are common in the literature. For example, assumed
density filtering, when the computed posterior is projected to simpler
distributions, has been suggested
\cite{boyen98tractable,minka01family,opper99bayesian}. We shall use the method of
conjugated priors \cite{gelman03bayesian} instead. For matrix $\b{A}$ we assume a
matrix valued normal distribution prior. For covariance matrix $\b{V}$ inverted
Wishart (IW) distribution will be our prior. One can show for these choices that the
functional form of the posteriori distributions is not affected.

We define the normally distributed matrix valued stochastic variable $\b{A} \in \R^{d\times m}$ by
using the following quantities: $\b{M} \in \R^{d\times m}$ is the expected value of $\b{A}$. $\b{V}
\in \R^{d\times d}$ is the covariance matrix of the rows, and $\b{K} \in \R^{m\times m}$ is the
so-called precision parameter matrix that we shall modify in accordance with the Bayesian update.
They are both positive semi-definite matrices. The density function of the stochastic variable
$\b{A}$ is defined as:
\begin{eqnarray*}
\N_{\b{A}}(\b{M},\b{V},\b{K})=\frac{|\b{K}|^{d/2}}{|2\pi\b{V}|^{m/2}}\exp(-\frac{1}{2}tr((\b{A}-\b{M})^T\b{V}^{-1}(\b{A}-\b{M})\b{K})),
\end{eqnarray*}
where $tr$, $|\cdot|$, and superscript $T$ denote the trace operation, the
determinant, and transposition, respectively. See e.g.
\cite{gupta99matrix,minka00bayesian}. We assume that $\b{Q} \in \R^{d \times d}$ is
a positive definite matrix and $n>0$. Using these notations, the density of the
Inverted Wishart distribution with parameters $\b{Q}$ and $n$ is as follows
\cite{gupta99matrix}:
\begin{eqnarray*}
\mathcal{IW}_{\b{V}}(\b{Q},n)=\frac{1}{Z_{n,d}}\frac{1}{|\b{V}|^{(d+1)/2}}\left|\frac{\b{V}^{-1}\b{Q}}{2}\right|^{n/2}
\exp(-\frac{1}{2}tr(\b{V}^{-1}\b{Q})),
\end{eqnarray*}
where $Z_{n,d}=\pi^{d(d-1)/4}\prod \limits_{i=1}^d \,
\Gamma((n+1-i)/2)$
 and $\Gamma(.)$ denotes the gamma function.

Now, one can rewrite model \eqref{e:simplified_model} as follows:
\begin{eqnarray}
P(\b{A}|\b{V})&=&\N_{\b{A}}(\b{M},\b{V},\b{K}), \label{e:bayesian_model1}\\
P(\b{V})&=&\mathcal{IW}_{\b{V}}(\b{Q},n), \label{e:bayesian_model2}\\
P(\b{e}_t|\b{V})&=&\N_{\b{e}_t}(\b{0},\b{V}), \label{e:bayesian_model3}\\
P(\b{y}_t|\b{A},\b{x}_t,\b{V})&=&\N_{\b{y}_t}(\b{Ax}_t,\b{V}).\label{e:bayesian_model4}
\end{eqnarray}

%%%%%%%%%%%%%%%%%%%%%%%%%%%%%%%%%%%%%%
\section{The Infomax Approach for Parameter Learning} \label{s:Infomax}
%%%%%%%%%%%%%%%%%%%%%%%%%%%%%%%%%%%%%%
Let us compute the parameter estimation strategy for task \eqref{e:Modell} (i.e., for task
\eqref{e:bayesian_model1}-\eqref{e:bayesian_model4}) as prescribed by the infomax principle. Let us
introduce two shorthands; $\bm{\theta}=\{\b{A},\b{V}\}$, and
$\{\b{x}\}_i^{j}=\{\b{x}_i,\ldots,\b{x}_j\}$. We choose the control value in \eqref{e:Modell} at
each instant such that it provides the most expected information concerning the unknown parameters.
Assuming that $\{\b{x}\}_1^{t}$, $\{\b{y}\}_1^{t}$ are given, according to the infomax principle
our goal is to compute
\begin{eqnarray}
\arg \max_{\b{u}_{t+1}}
I(\bm{\theta},\b{y}_{t+1};\{\b{x}\}_1^{t+1},\{\b{y}\}_1^{t}),
\label{e:goal1}
\end{eqnarray}
where $I(a,b;c)$ denotes the mutual information of stochastic
variables $a$ and $b$ for fixed parameters $c$. Let $H(a|b;c)$
denote the conditional entropy of variable $a$ conditioned on
variable $b$ and for fixed parameter $c$. Note that
\begin{eqnarray*}
I(\bm{\theta},\b{y}_{t+1};\{\b{x}\}_1^{t+1},\{\b{y}\}_1^{t})=
H(\bm{\theta};\{\b{x}\}_1^{t+1},\{\b{y}\}_1^{t})-H(\bm{\theta}|\b{y}_{t+1};
\{\b{x}\}_1^{t+1},\{\b{y}\}_1^{t}),
\end{eqnarray*}
holds \cite{cover91elements} and
$H(\bm{\theta};\{\b{x}\}_1^{t+1},\{\b{y}\}_1^{t})=H(\bm{\theta};\{\b{x}\}_1^{t},\{\b{y}\}_1^{t})$
is independent from $\b{u}_{t+1}$, hence our task is reduced to the evaluation of
the following quantity:
\begin{eqnarray}
\arg \min_{\b{u}_{t+1}}
H(\bm{\theta}|\b{y}_{t+1};\{\b{x}\}_1^{t+1},\{\b{y}\}_1^{t})=\hspace{5cm}
\label{e:Ent_cost}\\ =\arg \min_{\b{u}_{t+1}} -\int d \b{y}_{t+1}
P(\b{y}_{t+1}|\{\b{x}\}_1^{t+1},\{\b{y}\}_1^{t})\int
d\bm{\theta}P(\bm{\theta}|\{\b{x}\}_1^{t+1},\{\b{y}\}_1^{t+1})\log
P(\bm{\theta}|\{\b{x}\}_1^{t+1},\{\b{y}\}_1^{t+1}). \nonumber
\end{eqnarray}
\mbox{In order to solve this minimization problem we need to evaluate
$P(\b{y}_{t+1}|\{\b{x}\}_1^{t+1},\{\b{y}\}_1^{t})$}, the posterior \mbox{$
P(\bm{\theta}|\{\b{x}\}_1^{t+1},\{\b{y}\}_1^{t+1}) $}, and the entropy of the posterior, that is
\mbox{$\int d\bm{\theta}P(\bm{\theta}|\{\b{x}\}_1^{t+1},\{\b{y}\}_1^{t+1})\log
P(\bm{\theta}|\{\b{x}\}_1^{t+1},\{\b{y}\}_1^{t+1}) $}, where $P(a|b)$ denotes the conditional
probability of variable $a$ given condition $b$. The main steps of these computations are provided
below.

Assume that the \textit{a priori} distributions
$P(\b{A}|\b{V},\{\b{x}\}_1^{t},\{\b{y}\}_1^{t})=\N(\b{A}|\b{M}_t,\b{V},\b{K}_t)$
and
$P(\b{V}|\{\b{x}\}_1^{t},\{\b{y}\}_1^{t})=\IW_{\b{V}}(\b{Q}_t,n_t)$
are known. Then the posterior distribution of $\bm{\theta}$ is:
\begin{eqnarray*}
P(\b{A},\b{V}|\{\b{x}\}_1^{t+1},\{\b{y}\}_1^{t+1})&=&
\frac{P(\b{y}_{t+1}|\b{A},\b{V},\b{x}_{t+1})P(\b{A}|\b{V},\{\b{x}\}_1^{t},\{\b{y}\}_1^{t})P(\b{V}|\{\b{x}\}_1^{t},\{\b{y}\}_1^{t})}
{P(\b{y}_{t+1}|\{\b{x}\}_1^{t+1},\{\b{y}\}_1^{t})} \hspace{5mm}\nonumber \\
&=&\frac{\N_{\b{y}_{t+1}}(\b{Ax}_{t+1},\b{V})\N_{\b{A}}(\b{M}_{t},\b{V},\b{K}_{t})
\mathcal{IW}_{\b{V}}(\b{Q}_t,n_t)} {\int_{\b{A}} \int_{\b{V}}
\N_{\b{y}_{t+1}}(\b{Ax}_{t+1},\b{V})\N_{\b{A}}(\b{M}_{t},\b{V},\b{K}_{t})
\mathcal{IW}_{\b{V}}(\b{Q}_t,n_t))}\,.
\end{eqnarray*}
This expression can be rewritten in a more useful form: let $\b{K} \in \R^{m \times
m}$ and $\b{Q}\in \R^{d \times d}$ be positive definite matrices. Let $\b{A} \in
\R^{ d\times m}$, and let us introduce the density function of the matrix valued
Student-t distribution \cite{kotz04multivariate,minka00bayesian} as follows:
\begin{eqnarray*}
\mathcal{T}_{\b{A}}(\b{Q},n,\b{M},\b{K})=\frac{|\b{K}|^{d/2}}{\pi^{dm/2}}
\frac{Z_{n+m,d}}{Z_{n,d}}\frac{|\b{Q}|^{n/2}}{|\b{Q}+(\b{A}-\b{M})\b{K}(\b{A}-\b{M})^T|^{(m+n)/2}},
\end{eqnarray*}
Now, we need the following lemma:
\begin{lem} \label{l:transform1}
\begin{eqnarray}
\N_{\b{y}}(\b{A}\b{x},\b{V})\N_{\b{A}}(\b{M},\b{V},\b{K})\IW_{\b{V}}(\b{Q},n)
=  \nonumber
\N_\b{A}((\b{M}\b{K}+\b{y}\b{x}^T)(\b{x}\b{x}^T+\b{K})^{-1},\b{V},\b{x}\b{x}^T+\b{K})\times
\nonumber\\
\times\IW_{\b{V}}\left(\b{Q}+
\left(\b{y}-\b{Mx}\right)(1-\b{x}^T(\b{x}\b{x}^T+\b{K})^{-1}\b{x})\left(\b{y}-\b{Mx}\right)^T,n+1
\right)\times \nonumber\\
\times\T_{\b{y}}\left(\b{Q},n,\b{Mx},1-\b{x}^T(\b{x}\b{x}^T+\b{K})^{-1}\b{x}\right).
\nonumber
\end{eqnarray}

\end{lem}

\begin{proof}

%$\b{M}^+=(\b{M}\b{K}+\b{y}\b{x}^T)(\b{x}\b{x}^T+\b{K})^{-1} \in
%\R^{d \times m}$. Further, let $n^+=n + 1$, and let
%$\b{Q}^+=\b{Q}+(\b{A}-\b{M})\b{K}(\b{A}-\b{M})^T$.

It is easy to show that the following equations hold:
\begin{eqnarray}
\N_{\b{y}}(\b{A}\b{x},\b{V})\N_{\b{A}}(\b{M},\b{V},\b{K})&=&
\N_\b{A}(\b{M}^+,\b{V},\b{x}\b{x}^T+\b{K})\N_\b{y}(\b{M}\b{x},\b{V},\gamma), \nonumber\\
\N_{\b{A}}(\b{M},\b{V},\b{K})\IW_{\b{V}}(\b{Q},n)&=&
\IW_{\b{V}}\left(\b{Q}+\b{H},n+1\right)
\T_{\b{A}}(\b{Q},n,\b{M},\b{K}), \label{e:bayes_IW_T}
\end{eqnarray}
where $\b{M}^+=(\b{M}\b{K}+\b{y}\b{x}^T)(\b{x}\b{x}^T+\b{K})^{-1}$,
$\gamma=1-\b{x}^T(\b{x}\b{x}^T+\b{K})^{-1}\b{x}$, $\b{H}=(\b{A}-\b{M})\b{K}(\b{A}-\b{M})^T$ for the
sake of brevity. Then we have
\begin{eqnarray}
\N_\b{y}(\b{M}\b{x},\b{V},\gamma) \IW_{\b{V}}(\b{Q},n) =\IW_{\b{V}}(
\b{Q}+
\left(\b{y}-\b{Mx}\right)\gamma\left(\b{y}-\b{Mx}\right)^T,n+1)\T_{\b{y}}\left(\b{Q},n,\b{Mx},\gamma\right),\nonumber
\end{eqnarray}
and the statement of the lemma follows.
\end{proof}

Using this lemma, we can compute the posterior probabilities. Let us introduce the following
quantities:
\begin{eqnarray}
\gamma_{t+1}&=&1-\b{x}_{t+1}^T(\b{x}_{t+1}\b{x}_{t+1}^T+\b{K}_t)^{-1}\b{x}_{t+1}, \nonumber \\
n_{t+1}&=&n_t+1,  \nonumber\\
\b{M}_{t+1}&=&(\b{M}_{t}\b{K}_t+\b{y}_{t+1}\b{x}_{t+1}^T)(\b{x}_{t+1}\b{x}_{t+1}^T+\b{K}_t)^{-1},
\nonumber \\
\b{Q}_{t+1}&=&\b{Q}_t+\left(\b{y}_{t+1}-\b{M}_t\b{x}_{t+1}\right)
\gamma_{t+1}\left(\b{y}_{t+1}-\b{M}_t\b{x}_{t+1}\right)^T.
\label{e:Qupdate}
\end{eqnarray}
For the posterior probabilities we have determined that
\begin{eqnarray}
P(\b{A}|\b{V},\{\b{x}\}_1^{t+1},\{\b{y}\}_1^{t+1})&=&
\label{e:posteriorA}
\N_\b{A}(\b{M}_{t+1},\b{V},\b{x}_{t+1}\b{x}_{t+1}^T+\b{K}_{t}), \\
P(\b{V}|\{\b{x}\}_1^{t+1},\{\b{y}\}_1^{t+1})&=&\IW_{\b{V}}\left(\b{Q}_{t+1},n_{t+1}\right), \label{e:posteriorV}\\
P(\b{y}_{t+1}|\{\b{x}\}_1^{t+1},\{\b{y}\}_1^{t})
&=&\T_{\b{y}_{t+1}}\left(\b{Q}_t,n_t,\b{M}_t\b{x}_{t+1},\gamma_{t+1}\right).
\nonumber
\end{eqnarray}
Having done so, we can compute the entropy of the posterior distribution of
$\bm{\theta}=\{\b{A},\b{V}\}$ by means of the following lemma:
\begin{lem}\label{l:entropy1}
The entropy of a stochastic variable with density function
$P(\b{A},\b{V})=\N_{\b{A}}(\b{M},\b{V},\b{K})\IW_{\b{V}}(\b{Q},n)$
assumes the form $-\frac{d}{2}\ln |\b{K}|+(\frac{m+d+1}{2})\ln
|\b{Q}| + f_1(d,n)$, where $f_1(d,n)$ depends only on $d$ and $n$.
\end{lem}
\begin{proof}

%%%%%%%%%%%%%%%%%%%%%%%%%%%%%%%%%%%%%%%%%%%%%%%%
%\section{Proof of \ref{l:entropy1} lemma} \label{p:proof_entropy1}
%\section*{Proof of 4.2 lemma} \label{p:proof_entropy1}
%%%%%%%%%%%%%%%%%%%%%%%%%%%%%%%%%%%%%%%%%%%%%%%%
Let $vec(\b{A})$ denote a vector of $dm$ dimensions where the $(d(i-1)+1)^{th},
\ldots, (id)^{th}$ ($1\le i \le m$) elements of this vector are equal to the
elements of the $i^{th}$ column of matrix $\b{A} \in \R^{d \times m}$ in the
appropriate order. Let $\otimes$ denote the Kronecker-product. It is known that for
$P(\b{A})=\N_\b{A}(\b{M},\b{V},\b{K})$,
\mbox{$P(vec(\b{A}))=\N_{vec(\b{A})}(vec(\b{M}),\b{V}\otimes\b{K}^{-1})$} holds
\cite{minka00bayesian}. Using the well-known formula for the entropy of a
multivariate and normally distributed variable \cite{cover91elements} and applying
the relation \mbox{$|\b{V}\otimes\b{K}^{-1} |=|\b{V}|^m/|\b{K}|^d$}, we have that
\begin{eqnarray*}
H(\b{A};\b{V})=\frac{1}{2}\ln |\b{V}\otimes\b{K}^{-1}
|+\frac{dm}{2}\ln(2\pi e) =\frac{m}{2}\ln|\b{V}|-\frac{d}{2}\ln
|\b{K}|+\frac{dm}{2}\ln(2\pi e).
\end{eqnarray*}
Exploit certain properties of the Wishart distribution, we compute the entropy of distribution
$\IW_{\b{V}}(\b{Q},n)$. The density of the Wishart distribution is defined by
\begin{eqnarray*}
\mathcal{W}_{\b{V}}(\b{Q},n)=\frac{1}{Z_{n,d}}|\b{V}|^{(n-d-1)/2}\left|\frac{\b{Q}^{-1}}{2}\right|^{n/2}
\exp\left(-\frac{1}{2}tr(\b{V}\b{Q}^{-1})\right).
\end{eqnarray*}
Let $\Psi$ denote the digamma function, and let $f_2(d,n)=-\sum_{i=1}^d
\Psi(\frac{n+1-i}{2}))-d \ln 2$. Replacing $\b{V}^{-1}$ with $\b{S}$, we have for
the Jacobian that $ |\frac{d \b{V}}{d\b{S}}|=|\frac{d
\b{S}^{-1}}{d\b{S}}|=|\b{S}|^{-(d+1)}$ \cite{gupta99matrix}. To proceed we use that
$E_{\mathcal{W}_{\b{S}}(\b{Q},n)}\b{S}=n\b{Q}$, and
$E_{\mathcal{W}_{\b{S}}(\b{Q},n)}\ln|\b{S}|=\ln|\b{Q}|-f_2(d,n)$,
\cite{beal03variational} and substitute them into $E_{\IW_{\b{V}}(\b{Q},n)} \ln
|\b{V}|$, and $E_{\IW_{\b{V}}(\b{Q},n)} tr(\b{Q}\b{V}^{-1})$:
\begin{eqnarray}
E_{\IW_{\b{V}}(\b{Q},n)} \ln
|\b{V}|=\nonumber\\
=\int
\frac{1}{Z_{n,d}}\frac{1}{|\b{V}|^{(d+1)/2}}\left|\frac{\b{V}^{-1}\b{Q}}{2}\right|^{n/2}
\exp\left(-\frac{1}{2}tr(\b{V}^{-1}\b{Q})\right) \ln |\b{V}| d\b{V}\nonumber\\
=-\int
\frac{1}{Z_{n,d}}|\b{S}|^{(d+1)/2}\left|\frac{\b{S}\b{Q}}{2}\right|^{n/2}
\exp\left(-\frac{1}{2}tr(\b{S}\b{Q})\right) \ln |\b{S}| |\b{S}|^{-d-1}d\b{S}\nonumber\\
=-\int
\frac{1}{Z_{n,d}}|\b{S}|^{(n-d-1)/2}\left|\frac{\b{Q}}{2}\right|^{n/2}
\exp\left(-\frac{1}{2}tr(\b{S}\b{Q})\right) \ln |\b{S}| d\b{S}\nonumber\\
=-E_{\mathcal{W}_{\b{S}}(\b{Q}^{-1},n)} \ln |\b{S}| \nonumber\\
=\ln |\b{Q}|+f_2(d,n) \label{e:forentropy1}.
\end{eqnarray}

One can also show that
\begin{eqnarray}
E_{\IW_{\b{V}}(\b{Q},n)} tr(\b{Q}\b{V}^{-1})=\nonumber\\
=\int\frac{1}{Z_{n,d}}\frac{1}{|\b{V}|^{(d+1)/2}}\left|\frac{\b{V}^{-1}\b{Q}}{2}\right|^{n/2}
\exp\left(-\frac{1}{2}tr(\b{V}^{-1}\b{Q})\right) tr(\b{Q}\b{V}^{-1})d\b{V}\nonumber\\
=\int
\frac{1}{Z_{n,d}}|\b{S}|^{(d+1)/2}\left|\frac{\b{S}\b{Q}}{2}\right|^{n/2}
\exp\left(-\frac{1}{2}tr(\b{S}\b{Q})\right) tr(\b{Q}\b{S}) |\b{S}|^{-d-1}d\b{S},\nonumber\\
=\int
\frac{1}{Z_{n,d}}|\b{S}|^{(n-d-1)/2}\left|\frac{\b{Q}}{2}\right|^{n/2}
\exp\left(-\frac{1}{2}tr(\b{S}\b{Q})\right) tr(\b{Q}\b{S}) d\b{S},\nonumber\\
= E_{\mathcal{W}_{\b{S}}(\b{Q}^{-1},n) } tr(\b{Q}\b{S}),\nonumber\\
=tr(\b{Q}\b{Q}^{-1}n)=nd \label{e:forentropy2}
\end{eqnarray}
We calculate the entropy of stochastic variable $\b{V}$ with distribution $\IW_{\b{V}}(\b{Q},n)$.
It follows from Eq.~\eqref{e:forentropy1} and Eq.~\eqref{e:forentropy2} that
\begin{eqnarray*}
H(\b{V})=\\= -E_{\IW_{\b{V}}(\b{Q},n)}
\left[-\ln(Z_{n,d})+\frac{n}{2}\ln\left|\frac{\b{Q}}{2}\right|
-\frac{n+d+1}{2}\ln|\b{V}|-\frac{1}{2}tr(\b{V}^{-1}\b{Q})\right] \\
=\ln(Z_{n,d})-\frac{n}{2}\ln\left|\frac{\b{Q}}{2}\right|+\frac{n+d+1}{2}\left[
\ln |\b{Q}|-\sum_{i=1}^d \Psi(\frac{n+1-i}{2}))-d \ln 2
\right]+\frac{nd}{2}\\
=\frac{d+1}{2}\ln|\b{Q}|+f_3(d,n),
\end{eqnarray*}
where $f_3(d,n)$ depends only on $d$ and $n$.

Given the results above, we complete the computation of entropy $H(\b{A},\b{V})$ as follows:
\begin{eqnarray*}
H\left(\b{A},\b{V}\right)=H(\b{A}|\b{V})+H(\b{V})=H(\b{V})+\int
d{\b{V}}
\IW_{\b{V}}(\b{Q},n)H(\b{A};\b{V}) \\
=\int d{\b{V}} \IW_{\b{V}}(\b{Q},n)\left(\frac{m}{2} \ln
|\b{V}|-\frac{d}{2}\ln |\b{K}|+\frac{dm}{2}\ln(2\pi
e)\right)+H(\b{V}) \\
 =-\frac{d}{2}\ln |\b{K}|+\frac{dm}{2}\ln(2\pi e)+\frac{m}{2}
[\ln |\b{Q}|+f_2(d,n)]+\frac{d+1}{2}\ln|\b{Q}|+f_3(d,n)\\
=-\frac{d}{2}\ln |\b{K}|+(\frac{m+d+1}{2})\ln |\b{Q}| + f_1(d,n).
\end{eqnarray*}
This is exactly what was claimed in Lemma \ref{l:entropy1}.

\end{proof}

Lemmas \ref{l:transform1} and \ref{l:entropy1} lead to the
following:
\begin{cor}
For the entropy of a stochastic variable with posterior distribution $P(\b{A},\b{V}|\b{x},\b{y})$
it holds that
\begin{eqnarray}
H(\b{A},\b{V};\b{x},\b{y})=-\frac{d}{2}\ln |\b{x}\b{x}^T+\b{K}| +
f_1(d,n)+(\frac{m+d+1}{2})
\ln|\b{Q}+(\b{y}-\b{Mx})\gamma(\b{y}-\b{Mx})^T|. \nonumber
\end{eqnarray}
\end{cor}

We note that the following lemma also holds:
\begin{lem} \label{lem:invariance}
\begin{eqnarray*}\label{e:asdf}
\int \T_{\b{y}}\left(\b{Q},n,\bm{\mu},\gamma\right)\ln
|\b{Q}+(\b{y}-\bm{\mu}) \gamma(\b{y}-\bm{\mu})^T| d\b{y}
\end{eqnarray*}
is independent from both $\bm{\mu}$ and $\gamma$,
\end{lem}
and thus we can compute the conditional entropy expressed in \eqref{e:Ent_cost}:
\begin{lem} \label{l:entropy2}
\begin{eqnarray*}
H(\b{A},\b{V}|\b{y};\b{x})=\int p\left(\b{y}|\b{x}\right)
H(\b{A},\b{V};\b{x},\b{y}) d\b{y}= -\frac{d}{2}\ln
|\b{x}\b{x}^T+\b{K}|+g_1(\b{Q},d,n).
\end{eqnarray*}
where  $g_1(\b{Q},d,n)$ depends only on $\b{Q}$, $d$ and $n$.
\end{lem}
%%%%%%%%%%%%%%%%%%%%%%%%%%%%%%%%%%%%%%%%%%%%%%%%
%\section{Proof of \ref{l:entropy2} lemma} \label{p:proof_entropy2}
%\section*{Proof of 4.4 lemma} \label{p:proof_entropy2}
%%%%%%%%%%%%%%%%%%%%%%%%%%%%%%%%%%%%%%%%%%%%%%%%

%
%Let $\gamma=1-\b{x}^T(\b{x}\b{x}^T+\b{K})^{-1}\b{x}$. Then, as a
%result of Eq.~(\ref{e:asdf}), we have
%\begin{eqnarray}
%\int p\left(\b{y}|\b{x}\right) H(\b{A},\b{V}|\b{x},\b{y})
%d\b{y}=\int d{\b{y}}
%\prod_{i=1}^d\T_{y_i}\left(\beta_i,\alpha_i,(\b{Mx})_i,\frac{\gamma}{2}\right)\times
%\\
%\times \{ -\frac{d}{2}\ln
%|\b{x}\b{x}^T+\b{K}|+(1+\frac{m}{2})\sum_{i=1}^d\ln
%\{\beta_i+\frac{1}{2}[y_i-(\b{Mx})_i]^2\gamma\}+\sum_{i=1}^d
%f_1(\alpha_i+1/2) \} \nonumber\\
%=-\frac{d}{2}\ln |\b{x}\b{x}^T+\b{K}|+g(\bm{\alpha},\bm{\beta}),
%\nonumber
%\end{eqnarray}
%which proves the lemma.

Collecting all the terms, we arrive at the following \emph{intriguingly simple} expression
\begin{eqnarray}
\b{u}_{{t+1}^{opt}}&=&\arg\min_{\b{u}_{t+1}} \int
p\left(\b{y}_{t+1}|\{\b{x}\}_{1}^{t+1},\{\b{y}\}_{1}^{t}\right)
H(\b{A},\b{V}|\{\b{x}\}_{1}^{t+1},\{\b{y}\}_{1}^{t},\b{y}_{t+1}) d\b{y}_{t+1} \hspace{1cm}\nonumber \\
&=&\arg\min_{\b{u}_{t+1}} -\frac{d}{2}\ln
|\b{x}_{t+1}\b{x}_{t+1}^T+\b{K}_{t}|=\arg \max_{\b{u}_{t+1}}
\b{x}_{t+1}^T \b{K}_{t}^{-1}\b{x}_{t+1}, \label{e:cost_fun1}
\end{eqnarray}
where
\begin{eqnarray*}
\b{x}_{t+1}\doteq[\b{r}_{t-I};
\ldots;\b{r}_{t};\b{u}_{t-J+1};\ldots;\b{u}_{t+1}],
\end{eqnarray*}
and we used that \mbox{$|\b{x}\b{x}^T+\b{K}|=|\b{K}|(1+\b{x}^T\b{K}^{-1}\b{x})$}
according to the Matrix Determinant Lemma \cite{harville97matrix}. We assume a
bounded domain $\mathcal{U}$ for the control, which is necessary to keep the
maximization procedure of \eqref{e:cost_fun1} finite. This is, however, a reasonable
condition for all practical applications. So,
\begin{eqnarray}
\b{u}_{{t+1}^{opt}}&=&\arg \max_{\b{u} \in \mathcal{U}} \b{x}_{t+1}^T \b{K}_{t}^{-1}\b{x}_{t+1},
\label{e:cost_fun11}
\end{eqnarray}
In what follows D-optimal control will be referred to as \emph{`infomax interrogation scheme'}. The
steps of our algorithm are summarized in Table~\ref{tab:pseudocode}.

\begin{table}[h!]
  \centering
  \caption{Pseudocode of the algorithm} \label{tab:pseudocode}
  \begin{tabular}{|l|}
        \hline
        \textbf{Control Calculation}\\
        \verb|   |$\b{u}_{t+1}=\arg \max_{\b{u} \in \mathcal{U}} \b{\hat{x}}_{t+1}^T \b{K}_t^{-1}\b{\hat{x}}_{t+1}$\\
        \verb|   |where $\b{\hat{x}}_{t+1}=[\b{r}_{t-I}; \ldots;\b{r}_{t};\b{u}_{t-J+1};\ldots;\b{u}_{t};\b{u}]$\\
        \verb|   |set $\b{x}_{t+1}=[\b{r}_{t-I}; \ldots;\b{r}_{t};\b{u}_{t-J+1};\ldots;\b{u}_{t};\b{u_{t+1}}]$\\
        \textbf{Observation}\\
        \verb|   |observe $\b{r}_{t+1}$, and let $\b{y}_{t+1}=g^{-1}(\b{r}_{t+1})$\\
        \textbf{Bayesian update}\\
        \verb|   |$ \b{M}_{t+1}=(\b{M}_{t}\b{K}_{t}+\b{y}_{t+1}\b{x}_{t+1}^T)(\b{x}_{t+1}\b{x}_{t+1}^T+\b{K}_{t})^{-1}$ \\
        \verb|   |$ \b{K}_{t+1}=\b{x}_{t+1}\b{x}_{t+1}^T+\b{K}_{t}$ \\
        \verb|   |$n_{t+1}=n_{t}+1$ \\
        \verb|   |$ \bm{\gamma}_{t+1}=1-\b{x}^T_{t+1}(\b{x}_{t+1}\b{x}_{t+1}^T+\b{K}_{t})^{-1}\b{x}_{t+1}$ \\
        \verb|   |$ \b{Q}_{t+1}= \bm{Q}_{t}+\left(\b{y}_{t+1}-\b{M}_t\b{x}_{t+1}\right) \gamma_{t+1} \left(\b{y}_{t+1}-\b{M}_{t}\b{x}_{t+1}\right)^T$ \\
        \hline
 \end{tabular}
\end{table}

Computation of the inverse $(\b{x}_{t+1}\b{x}_{t+1}^T+\b{K}_{t})^{-1}$ in Table
\ref{tab:pseudocode} can be simplified considerably by the following recursion: let
$\b{P}_t=\b{K}_t^{-1}$, then according to the Sherman-Morrison formula
\cite{golub96matrix}
\begin{eqnarray}\label{eq:simple}
\b{P}_{t+1}=(\b{x}_{t+1}\b{x}_{t+1}^T+\b{K}_t)^{-1}=
\b{P}_{t}-\frac{\b{P}_{t}\b{x}_{t+1}\b{x}_{t+1}^T\b{P}_{t}}{1+\b{x}_{t+1}^T\b{P}_{t}\b{x}_{t+1}},
\end{eqnarray}
In this expression matrix inversion disappears and instead only a real number is inverted.

%%%%%%%%%%%%%%%%%%%%%%%%%%%%%%%%%%
\section{Estimating the Noise} \label{s:Estimating}
%%%%%%%%%%%%%%%%%%%%%%%%%%%%%%%%%%%%
One might wish to compute the optimal control for estimating noise
$\b{e}_t$ in \eqref{e:Modell}, instead of the identification problem
above. Based on \eqref{e:Modell} and because
\begin{eqnarray}
\b{e}_{t+1}=\b{y}_{t+1}-\sum_{i=0}^I \b{F}_i\b{r}_{t-i}-\sum_{j=0}^J
\label{e:noise} \b{B}_j\b{u}_{t+1-j},
\end{eqnarray}
one might think that the best strategy is to use the optimal infomax
control of Table~\ref{tab:pseudocode}, since it provides good
estimations for parameters
$\b{A}=[\b{F}_I,\ldots,\b{F}_0,\b{B}_{J},\ldots, \b{B}_{0}] $ and so
for noise $\b{e}_t$.

Another---and different---thought is the following. At time $t$, let us denote our estimations as
$\hat{\b{e}}_t$, $\hat{\b{F}}_i^t$ (i=0,\ldots,I), and $\hat{\b{B}}_j^t$ (j=0,\ldots,J). Using
\eqref{e:noise}, we have that
\begin{eqnarray}\label{e:A}
\b{e}_{t+1}-\hat{\b{e}}_{t+1}=\sum_{i=0}^I
(\b{F}_i-\hat{\b{F}}_i^t)\b{r}_{t-i}+\sum_{j=0}^J(\b{B}_j-\hat{\b{B}}_j^t)\b{u}_{t+1-j}.
\end{eqnarray}
This hints that the control should be $\b{u}_{t}=\b{0}$ for all
times in order to get rid of the error contribution of matrix
$\b{B}_j$ in \eqref{e:A}.

Straightforward utilization of D-optimality considerations, opposed to the objective of
\eqref{e:goal1}, suggests the optimization of the following quantity:
\begin{eqnarray*}
\arg \max_{\b{u}_{t+1}}
I(\b{e}_{t+1},\b{y}_{t+1};\{\b{x}\}_1^{t+1},\{\b{y}\}_1^{t}),
\label{e:goal2}
\end{eqnarray*}
That is, for the estimation of the noise we want to design a control signal $\b{u}_{t+1}$ such that
the next output is the best from the point of view of greedy optimization of mutual information
between the next output $\b{y}_{t+1}$ and the noise $\b{e}_{t+1}$. It is easy to show that this
task is equivalent to the following optimization problem:
\begin{eqnarray}
\arg \min_{\b{u}_{t+1}} \int d \b{y}_{t+1} P(\b{y}_{t+1}|\{\b{x}\}_1^{t+1},\{\b{y}\}_1^{t})
H(\b{e}_{t+1};\{\b{x}\}_1^{t+1},\{\b{y}\}_1^{t+1}), \label{e:noise_integral}
\end{eqnarray}
where $H(\b{e}_{t+1};\{\b{x}\}_1^{t+1},\{\b{y}\}_1^{t+1})=
H(\b{A}\b{x}_{t+1};\{\b{x}\}_1^{t+1},\{\b{y}\}_1^{t+1})$, because
$\b{e}_{t+1}=\b{y}_{t+1}-\b{A}\b{x}_{t+1}$. To compute this quantity we need the
following lemma \cite{minka00bayesian}:
\begin{lem}
If $P(\b{A})=\N_{\b{A}}(\b{M},\b{V},\b{K})$, then
$P(\b{Ax})=\N_{\b{Ax}}\left(\b{Mx},\b{V},\left(\b{x}^T\b{K}^{-1}\b{x}\right)^{-1}\right)$
\end{lem}
Applying this lemma and using \eqref{e:posteriorA} one has that
\begin{eqnarray}
P(\b{A}\b{x}_{t+1}|\b{V},\{\b{x}\}_1^{t+1},\{\b{y}\}_1^{t})=\N_{\b{A}\b{x}_{t+1}}\left(\b{M}_{t+1}\b{x}_{t+1},\b{V},
\left(\b{x}_{t+1}^T\b{K}_{t+1}^{-1}\b{x}_{t+1}\right)^{-1}\right)
\label{e:Axposterior}
\end{eqnarray}
We introduce the notations
\begin{eqnarray}
\tilde{K}_{t+1}&=&\left(\b{x}_{t+1}^T\b{K}_{t+1}^{-1}\b{x}_{t+1}\right)^{-1}
\in \R, \label{e:Ktilde}\\
\lambda_{t+1}&=&1+(\b{A}\b{x}_{t+1}-\b{M}_{t+1}\b{x}_{t+1})^T(\tilde{K}_{t+1}\b{Q}^{-1}_{t+1})
(\b{A}\b{x}_{t+1}-\b{M}_{t+1}\b{x}_{t+1}) \in \R \nonumber
\end{eqnarray}
and use \eqref{e:bayes_IW_T} and \eqref{e:posteriorV} for the posterior distribution
\eqref{e:Axposterior}. Then we arrive at
\begin{eqnarray*}
P(\b{A}\b{x}_{t+1}|\{\b{x}\}_1^{t+1},\{\b{y}\}_1^{t})&=&
\T_{\b{A}\b{x}_{t+1}}
\left(\b{Q}_{t+1},n_{t+1},\b{M}_{t+1}\b{x}_{t+1},\tilde{K}_{t+1}\right)\\
&=&\pi^{-d/2}|\tilde{K}^{-1}_{t+1}\b{Q}_{t+1}|^{-1/2}\frac{\Gamma(\frac{n_{t+1}+1}{2})}{\Gamma(\frac{n_{t+1}+1-d}{2})}\lambda_{t+1}^{\frac{n_{t+1}+1}{2}}
\end{eqnarray*}
The Shannon-entropy of this distribution according to \cite{zografos05expressions}
equals:
\begin{eqnarray*}
H(\b{A}\b{x}_{t+1};\{\b{x}\}_1^{t+1},\{\b{y}\}_1^{t+1})&=&
f_4(d,n_{t+1})+\frac{d}{2}\log|\tilde{K}^{-1}_{t+1}|+\log|\b{Q}_{t+1}|
\end{eqnarray*}
 where
$$f_4(d,n_{t+1})=-\log\frac{\Gamma(\frac{n_{t+1}+1}{2})}{\pi^{d/2}\Gamma(\frac{n_{t+1}+1-d}{2})}+
\frac{n_{t+1}+1}{2}\left(\Psi\left(\frac{n_{t+1}+1}{2}\right)-
\Psi\left(\frac{n_{t+1}+1-d}{2}\right)\right).$$ Using the notations introduced in
\eqref{e:Qupdate} and in \eqref{e:Ktilde} the above expressions can be transcribed as follows:
\begin{eqnarray*}
H(\b{A}\b{x}_{t+1};\{\b{x}\}_1^{t+1},\{\b{y}\}_1^{t+1})
&=& f_4(d,n_{t+1})-\frac{d}{2}\log|\tilde{K}_{t+1}|+\log|\b{Q}_{t+1}|\\
&=&f_4(d,n_{t+1})+\frac{d}{2}\log|\b{x}_{t+1}^T(\b{K}_{t}+\b{x}_{t+1}\b{x}_{t+1}^T)^{-1}\b{x}_{t+1}|+\log|\b{Q}_{t+1}|\\
&=&f_4(d,n_{t+1})+\frac{d}{2}\log|\b{x}_{t+1}^T(\b{K}_{t}+\b{x}_{t+1}\b{x}_{t+1}^T)^{-1}\b{x}_{t+1}|+\\
&&+\log|\b{Q}_t+\left(\b{y}_{t+1}-\b{M}_t\b{x}_{t+1}\right)
\gamma_{t+1}\left(\b{y}_{t+1}-\b{M}_t\b{x}_{t+1}\right)^T |
\end{eqnarray*}
Now, we are in a position to calculate \eqref{e:noise_integral} by applying Lemma
\ref{lem:invariance} as before. We get that
\begin{eqnarray*}
\int d \b{y}_{t+1} P(\b{y}_{t+1}|\{\b{x}\}_1^{t+1},\{\b{y}\}_1^{t})
H(\b{e}_{t+1};\{\b{x}\}_1^{t+1},\{\b{y}\}_1^{t+1})=\\
=g_2(\b{Q}_t,d,n_{t+1})+\frac{d}{2}\log|\b{x}_{t+1}^T(\b{K}_{t}+\b{x}_{t+1}\b{x}_{t+1}^T)^{-1}\b{x}_{t+1}|,
\end{eqnarray*}
where $g_2(\b{Q}_t,d,n_{t+1})$ depends only on $\b{Q}_t$, $d$ and
$n_{t+1}$. Thus, we have that
\begin{eqnarray*}
\arg \max_{\b{u}_{t+1}}
I(\b{e}_{t+1},\b{y}_{t+1};\{\b{x}\}_1^{t+1},\{\b{y}\}_1^{t})=\arg
\min_{\b{u}_{t+1}}
\log|\b{x}_{t+1}^T(\b{K}_{t}+\b{x}_{t+1}\b{x}_{t+1}^T)^{-1}\b{x}_{t+1}|\\
=\arg \min_{\b{u}_{t+1}}
\log\left|\b{x}_{t+1}^T\left(\b{K}_{t}^{-1}-\frac{\b{K}_{t}^{-1}\b{x}_{t+1}\b{x}_{t+1}^T\b{K}_{t}^{-1}}
{1+\b{x}_{t+1}^T\b{K}_{t}^{-1}\b{x}_{t+1}} \right)\b{x}_{t+1}\right|\\
=\arg \min_{\b{u}_{t+1}}
\log\left| \frac{\b{x}_{t+1}^T\b{K}_{t}^{-1}\b{x}_{t+1}}{1+\b{x}_{t+1}^T\b{K}_{t}^{-1}\b{x}_{t+1}}\right|\\
=\arg \min_{\b{u}_{t+1}} \b{x}_{t+1}^T \b{K}_t^{-1}\b{x}_{t+1}
\end{eqnarray*}

In practice, we perform this optimization in an appropriate domain $\mathcal{U}$. Thus, the
D-optimal interrogation scheme for noise estimation is as follows
\begin{eqnarray}
\b{u}_{t+1}^{opt}=\arg \min_{\b{u} \in \mathcal{U}} \b{x}_{t+1}^T
\b{K}_t^{-1}\b{x}_{t+1}. \label{e:error_control}
\end{eqnarray}

%\new{As opposed to the infomax control of \eqref{e:cost_fun11} we will call
%the interrogation \eqref{e:error_control} \emph{infomax noise}
%control.}

It is worth noting that this D-optimal cost function for noise
estimation and the D-optimal cost function derived for parameter
estimation in \eqref{e:cost_fun1} are not compatible with each
other. Estimating one of them quickly will necessarily delay the
estimation of the other.

In Section~\ref{s:greedy_non_greedy} we see that for large enough $t$ values, expression
\eqref{e:error_control} gives rise to control values close to $\b{u}_{t}=\b{0}$.

\subsection{Greedy and Non-Greedy Optimization}
\label{s:greedy_non_greedy}

Greedy optimization of \eqref{e:error_control} is simple, provided
that $\b{K}_t$ is fixed during the optimization of $\b{u}_{t+1}$. If
so, then the optimization task is quadratic. To see this, let us
partition matrix $\b{K}_t$ as follows:
\begin{eqnarray*}
\b{K}_t=\begin{pmatrix}
          \b{K}_t^{11} & \b{K}_t^{12} \\
          \b{K}_t^{21} & \b{K}_t^{22} \\
        \end{pmatrix},
\end{eqnarray*}
where $\b{K}_t^{11} \in \R^{d \times d}$,$\b{K}_t^{21} \in \R^{m-d \times d}$, $\b{K}_t^{22} \in
\R^{m-d \times m-d}$. It is easy to see that if domain $\mathcal{U}$ in \eqref{e:error_control} is
large enough  then
\begin{eqnarray}
\b{u}_{t+1}^{opt}=(\b{K}_t^{22})^{-1}\b{K}_t^{21}\b{r}_t.
\label{e:error_control2}
\end{eqnarray}

However, for non-greedy solutions, expression $\b{K}_t$ in $\b{x}_{t+1}^T \b{K}_t^{-1}\b{x}_{t+1}$
changes, because it may depend on previous control inputs $\b{u}_1,\ldots, \b{u}_t$, the subject of
previous optimization steps. The optimal strategy for long-term non-greedy optimization falls
outside of the scope of the present work. Here we propose the following heuristics for this
problem: Use the strategy of Table~\ref{tab:pseudocode} for the first $\tau$ steps. It increases
$|\b{K}_t|$ quickly in \eqref{e:error_control}. Then after $\tau$-steps switch to the control
described in \eqref{e:error_control2}. This will decrease the cost function \eqref{e:error_control}
further. We will call this non-greedy interrogation heuristics introduced for noise estimation
\emph{`$\tau$-infomax noise interrogation'}.

It is worth noting that in the $\tau$-infomax noise interrogation,
if the $\tau$ switching time is large enough then for large $t$
values $|\b{K}_t^{22}|$ will be large, and hence
---according to \eqref{e:error_control2}--- the optimal $\b{u}_t$ interrogation will be close to $\b{0}$.
The approximation of the `$\tau$-infomax noise interrogation' when we use the interrogation
described in Table \ref{tab:pseudocode} for $\tau$ steps and then switch to
\emph{zero-interrogation} will be called the \emph{`$\tau$-zero interrogation'} scheme.

%%%%%%%%%%%%%%%%%%%%%%%%%%%%%%%%%%%%%%%%%%%%%%%%%
\section{Discussion and Conclusions} \label{s:Discussion_and_Conclusions}
%%%%%%%%%%%%%%%%%%%%%%%%%%%%%%%%%%%%%%%%%%%%%%%%%

We have treated the identification problem of recurrent neural networks described by
model \eqref{e:Modell}. We applied active learning to solve this task. In
particular, the online D-optimality principle was applied and we investigated the
learning properties for parameter and noise estimations. We note that the D-optimal
interrogation scheme is also called infomax control in the literature
\cite{lewi07realtime}. This name originates from the cost function that optimizes
the mutual information.

The GLM model used by \cite{lewi07realtime} is as follows:
\begin{eqnarray} \b{r}_{t+1}=g\left(\sum_{i=0}^I
\b{F}_i\b{r}_{t-i}+\sum_{j=0}^J \b{B}_j\b{u}_{t+1-j}\right)+ \b{e}_{t+1}, \label{e:model_lewi}
\end{eqnarray}
where $\{\b{e}_t\}$ is i.i.d. noise with $\b{0} \in \R^d$ mean. The authors model spiking neurons
and assume that the main source of the noise is this spiking, which appears at the output of the
neurons and adds linearly to the neural activity. They investigated the case in which the observed
quantity $\b{r}_t$ had a Poisson distribution. Unfortunately, in this model Bayesian equations
become intractable and the estimation of the posterior may be spoiled, because the distribution is
projected to the family of normal distributions at each instant. A serious problem with this
approach is that the extent of the information loss caused by this approximation is not known. Our
stochastic RNN model
\begin{eqnarray*}
\b{r}_{t+1}&=&g\left(\sum_{i=0}^I \b{F}_i\b{r}_{t-i}+\sum_{j=0}^J \b{B}_j\b{u}_{t+1-j}+
\b{e}_{t+1}\right),
\end{eqnarray*}
differs only slightly from the GLM model of \eqref{e:model_lewi}, but it has considerable
advantages, as we shall discuss below.

Bayesian designs of different kinds were derived for the linear regression problem
in \cite{verdinelli00note}:
\begin{eqnarray}
\b{y}=\b{X}\bm{\theta}+\b{e} \label{e:Verdinelli1}\\
P(\b{e})=\N_{\b{e}}(0,\sigma^2\b{I}). \label{e:Verdinelli2}
\end{eqnarray}
This problem is similar to ours
(\eqref{e:bayesian_model1}-\eqref{e:bayesian_model4}), but while the goal of
\cite{verdinelli00note} was to find an optimal design for the explanatory variables
$\bm{\theta}$, we were concerned with the parameter ($\b{X}$ in
\eqref{e:Verdinelli1}) and the noise ($\b{e}$) estimation task. In Verdinelli's
paper inverted gamma prior and vector-valued normal distribution were assumed on the
isotropic noise and on the explanatory variables, respectively. By contrast, we were
interested in the matrix-valued coefficients and in general, non-isotropic noises.
We used matrix-valued normal distribution for the coefficients and inverted Wishart
distribution for the covariance matrix as conjugate priors. Due to the inverted
Wishart distribution that we used, the covariance matrix of the noise is not
restricted to the isotropic form, but can be general in our case.

The Bayesian online learning framework allowed us to derive analytic results for the greedy
optimization of the parameters as well as the driving noise. Optimal interrogation strategies
\eqref{e:cost_fun11} and \eqref{e:error_control} appeared in attractive, intriguingly simple
quadratic forms. We have shown that these two tasks are incompatible with each other. Parameter and
noise estimations require the maximization and the minimization of expression $\b{x}_{t+1}^T
\b{K}_{t}^{-1}\b{x}_{t+1}$, respectively.

The problem of non-greedy optimization of the full task has been left open. However,
we put forth a heuristic solution for the estimation of the driving noise that we
called $\tau$-infomax noise interrogation. It uses the D-optimal interrogation of
Table~\ref{tab:pseudocode} up to $\tau$-steps, and applies the noise estimation
control of \eqref{e:error_control} afterwards. This heuristics decreases the
estimation error of the coefficients of matrices $\b{F}$ and $\b{B}$ up to time
$\tau$ and thus --- upon turning off the explorative D-optimization --- tries to
minimize the estimation error of the value of the noise at time $\tau+1$. We
introduced the $\tau$-zero interrogation scheme and showed that it is a good
approximation of the $\tau$-infomax noise scheme for large $\tau$ values.

Finally, it seems desirable to determine the conditions under which the algorithms derived from the
D-optimal (infomax) principle are both consistent and efficient. The tractable form of our
approximation-free results is promising in this respect.

%%%%%%%%%%%%%%%%%%%%%%%%%%%%%%%%%%%%%%%%%%%%%%%%%
\section{Acknowledgments} \label{s:acknow}
%%%%%%%%%%%%%%%%%%%%%%%%%%%%%%%%%%%%%%%%%%%%%%%%%

This research has been supported by the EC NEST `Perceptual Consciousness:
Explication and Testing' grant under contract 043261. Opinions and errors in this
manuscript are the author's responsibility, they do not necessarily reflect the
opinions of the EC or other project members.

%\bibliographystyle{splncs}
%\bibliography{Infomax_Control}
\end{document}